\documentclass{article}
\usepackage{spconf,amsmath,graphicx} 
\usepackage[table,xcdraw]{xcolor} 
\usepackage{amsmath} 
\usepackage{booktabs} 
\usepackage{amssymb} 
\usepackage{enumitem} 
\usepackage{hyperref}
\definecolor{lightblue}{rgb}{0.678, 0.847, 0.902} 

\title{DEPTH-AWARE SCORING AND HIERARCHICAL ALIGNMENT 
\\FOR MULTIPLE OBJECT TRACKING}
\name{Milad Khanchi, Maria Amer, and Charalambos Poullis}
\address{Concordia University, Montreal, Quebec, Canada}
\begin{document}
%
\maketitle
\begin{abstract}
Current motion-based multiple object tracking (MOT) approaches rely heavily on Intersection-over-Union (IoU) for object association. Without using 3D features, they are ineffective in scenarios with occlusions or visually similar objects. To address this, our paper presents a novel depth-aware framework for MOT. We estimate depth using a zero-shot approach and incorporate it as an independent feature in the association process. Additionally, we introduce a Hierarchical Alignment Score that refines IoU by integrating both coarse bounding box overlap and fine-grained (pixel-level) alignment to improve association accuracy without requiring additional learnable parameters. To our knowledge, this is the first MOT framework to incorporate 3D features (monocular depth) as an independent decision matrix in the association step. Our framework achieves state-of-the-art results on challenging benchmarks without any training nor fine-tuning. The code is available at \href{https://github.com/Milad-Khanchi/DepthMOT}{https://github.com/Milad-Khanchi/DepthMOT}.
\end{abstract}
\vspace*{-0.4cm}
\section{Introduction}
\vspace*{-0.2cm}
\label{sec:intro}
Multiple object tracking (MOT)~\cite{vaquero2024lost, lv2024diffmot, shim2024confidence} involves detecting objects in video frames and continuously tracking them across time, requiring the simultaneous resolution of object detection, data association, and trajectory prediction. Challenges include noisy observations, occlusion, rapid motion, or similar objects. Recent research has focused on hybrid models that integrate both motion and appearance-based features. However, even with appearance cues, these approaches remain limited in scenarios that require spatial differentiation. For instance, two visually similar objects at different distances from the camera may appear indistinguishable in both motion and appearance, leading to frequent tracking errors. Moreover, occlusion scenarios often disrupt trajectory continuity, as traditional 2D Intersection over Union (IoU) based association is ineffective when objects overlap or move along similar paths. Furthermore, appearance-based approaches only extract appearance from cropped bounding boxes, neglecting the spatial context within each frame.

To address these challenges, we propose a novel depth-aware MOT framework that integrates monocular depth estimation into the tracking pipeline, allowing our framework to leverage 3D spatial cues for more robust association. By introducing zero-shot depth estimation, our framework differentiates between objects based on their distance from the camera, providing a strong discriminative feature for association. Our depth-aware tracking model is further enhanced by a new Hierarchical Alignment Score (HAS) for association, which combines bounding box IoU with pixel-level alignment, capturing both coarse and fine object alignment features to improve matching accuracy. Unlike conventional IoU-based approaches, HAS adaptively refines the alignment between objects by emphasizing shape similarity in scenarios where bounding box overlap alone may be insufficient.

To our knowledge, this is the first MOT framework to integrate monocular depth as an independent factor in the object association process for MOT.
Our contributions are:
%
\vspace*{-0.2cm}
\begin{itemize}[leftmargin=*]
    \setlength\itemsep{-0.15cm}
    \item We introduce a depth-aware tracking framework that incorporates zero-shot monocular depth estimation, providing robust spatial differentiation between objects based on their distance from the camera, thus improving association in complex scenarios.
    \item We propose a Hierarchical Alignment Score (HAS), a novel score that combines bounding box IoU with pixel-level alignment, enabling our framework to achieve precise object matching in cluttered and occluded environments.
    \item We present a comprehensive evaluation on challenging MOT benchmarks, showing that our approach resolves ambiguities in scenarios, such as occlusions and visually similar objects, without any training nor fine-tuning.
\end{itemize}
\vspace*{-0.1cm}
\vspace*{-0.4cm}
\section{RELATED WORKS}
\vspace*{-0.2cm}
\label{sec:RELATEDWORKS}
MOT approaches can be broadly categorized into: joint detection-ReIdentification (JDR)~\cite{zhang2021fairmot, xu2022transcenter} and tracking-by-detection (TBD)~\cite{vaquero2024lost, lv2024diffmot, cao2023observation, maggiolino2023deep, meinhardt2022trackformer}. Tracking can be performed frame-wise~\cite{lv2024diffmot, cao2023observation, maggiolino2023deep, meinhardt2022trackformer}, where the model processes each frame sequentially, utilizing information from previous frames. Alternatively, it can be conducted at the sequence level~\cite{braso2020learning, cetintas2023unifying, braso2022multi}, where the entire sequence is available during the association process. Our approach is frame-wise tracking-by-detection; we review related state-of-the-arts next.
\vspace*{-0.15cm}
\begin{itemize}[leftmargin=*]
    \setlength\itemsep{-0.15cm}
    \item \textbf{Tracking by Attention:} Attention-based methods leverage the Transformer architecture to associate objects. TrackFormer~\cite{meinhardt2022trackformer} integrates detection and tracking in an end-to-end attention-based model, further refined to MOTRv2 by Zhang et al.~\cite{zhang2023motrv2} with YOLOX detector~\cite{zheng2021yolox}. Gao et al.~\cite{gao2023memotr} implemented an end-to-end long-term memory-augmented transformer for MOT to refine detection and association.
    \item \textbf{Tracking by Regression:} Many frame-wise tracking models apply regression to associate detected objects across frames~\cite{shim2024confidence, lv2024diffmot, cao2023observation, maggiolino2023deep}. Kalman Filter-based approaches, such as OC-SORT~\cite{cao2023observation}, primarily leverage motion cues, while advanced models like Deep OC-SORT~\cite{maggiolino2023deep} and DiffMOT~\cite{lv2024diffmot} enhance the association process with appearance cues to improve tracking in dense scenes. Huang et al.~\cite{huang2024deconfusetrack} addressed confusion issues in MOT by proposing an association method and a detection post-processing technique.
    \item \textbf{Tracking by Depth:} Quach et al.~\cite{quach2024depth} incorporate dynamic control variables into a Kalman Filter, updating object states based on relative depth—defined as the depth ordering of detected objects w.r.t. the camera. However, their evaluation focused on the accuracy of depth estimation against RGB-D ground truth without demonstrating its direct impact on tracking performance. Wang et al.~\cite{wang2024robust} utilize depth from stereo cameras alongside camera pose data. This approach relies heavily on accurate camera intrinsics and pose calibration. Liu et al.~\cite{liu2022det} propose a depth-aware tracking model, yet limited to indoor tracking.
\end{itemize}
\vspace*{-0.2cm}
\vspace*{-0.4cm}
\section{Methodology}
\vspace*{-0.3cm}
\label{sec:METHOD}
\begin{figure*}[ht!]
  \centering
  \resizebox{\textwidth}{!}{\includegraphics{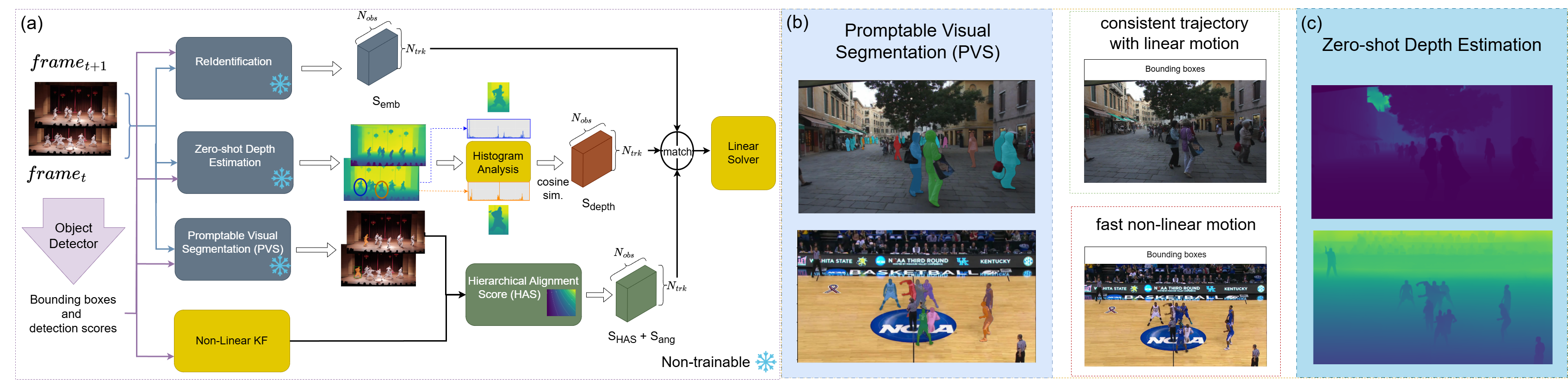}}
  \vspace*{-0.7cm}
  \caption{(a): Overview of the proposed framework, which integrates appearance scores from ReIdentification (RE-ID), motion scores derived from HAS, and depth scores. We use an advanced linear solver module and incorporate a PVS module for precise motion analysis. The Histogram Analysis block, as well as the RE-ID and HAS blocks, generate individual score matrices \( R \in \mathbb{R}^{N_{obs} \times N_{trk}} \), where \( N_{obs} \) represents the number of new observations in the current frame and \( N_{trk} \) denotes the number of tracklets. These score matrices capture similarities based on appearance, motion, and depth, enabling a comprehensive assessment for object association. The Histogram Analysis block, in particular, also performs a comparison of pixel intensity distributions between the depth maps of two objects within the same frame, highlighting variations in depth based on their distances from the camera. (b), (c): Examples. Zero-shot monocular depth estimation and PVS modules in two scenarios.}
  \label{fig:model_architecture}
\end{figure*}

In a TBD paradigm, objects are initially detected in individual frames and subsequently associated across time in a frame-wise manner. For object detection, we use the known model YOLOX~\cite{zheng2021yolox}. We perform object association in two steps. In the first step, we categorize detected objects based on their detection confidence (\(DF\)) as in~\cite{zhang2022bytetrack}: high-confidence (\(DF > 0.6\)) and low-confidence (\(DF \leq 0.6\)).  We match high-confidence objects with existing tracklets from previous frames using our matching technique detailed in Sec.~\ref{sec:matching}. In the second step, the remaining unmatched objects are associated with future positions of remaining tracklets, solely based on the 2D bounding box overlap (IoU). We predict the position of future tracklets using a non-linear Kalman filter (see Sec.~\ref{sec:appearance_motion_fusion}).

MOT frameworks heavily rely on IoU and 2D appearance-based ReIdentification (RE-ID) models, which may fail under heavy occlusion or similar objects. To address these limitations, our framework consists of four main components, as in Fig.~\ref{fig:model_architecture} (a): an appearance-motion fusion model (Sec.~\ref{sec:appearance_motion_fusion}), a depth-aware association process (Sec.~\ref{sec:depth}), and two modules for refining motion-based object associations — Promptable Visual Segmentation (PVS) (Sec.~\ref{sec:pvsseg}) and a novel Hierarchical Alignment Score (HAS) (Sec.~\ref{sec:HAS}). Notably, we avoid training or fine-tuning any components, relying on existing pre-trained models to ensure the generalization and adaptability of our framework across various scenarios.

\vspace*{-0.4cm}
\subsection{Appearance-Motion Fusion for MOT}
\vspace*{-0.2cm}
\label{sec:appearance_motion_fusion}
%
%
\textbf{Non-linear Kalman Filter:} Our DepthMOT framework incorporates non-linear Kalman Filter (KF) to create motion scores. KF-based models operate in two main steps: prediction and update.

In the \textbf{Prediction Step}, the state estimate \(x\) at time \(t\) is derived based on the state estimate at \(t-1\) using the transition model: 
\setlength{\abovedisplayskip}{5pt}   
\setlength{\belowdisplayskip}{5pt}   
\begin{equation}
\label{eq:kfpred}
\begin{aligned}
    \hat{x}_{t|t-1} &= \mathbf{F}_t \hat{x}_{t-1|t-1}, \\
    \mathbf{P}_{t|t-1} &= \mathbf{F}_t \mathbf{P}_{t-1|t-1} \mathbf{F}_t^\top + \mathbf{Q}_t,
\end{aligned}
\end{equation}
where \(\mathbf{F}_t\) is the state transition matrix, representing state evolution over time. \(\mathbf{P}_{t|t-1}\) is the error covariance matrix, captures the uncertainty of the predicted state. \(\mathbf{P}_{t|t-1}\) is updated by incorporating the process noise covariance matrix \(\mathbf{Q}_t\), accounting for model uncertainties.
%
%
In the \textbf{Update Step}, the Kalman Gain \(\mathbf{K}_t\) is computed to balance the predicted state against the observation as follows:
\setlength{\abovedisplayskip}{5pt}   
\setlength{\belowdisplayskip}{5pt}   
\begin{equation}
\label{eq:kfupdate}
\begin{aligned}
    \mathbf{K}_t &= \mathbf{P}_{t|t-1} \mathbf{H}_t^\top \left( \mathbf{H}_t \mathbf{P}_{t|t-1} \mathbf{H}_t^\top + \mathbf{R}_t \right)^{-1}, \\
    \hat{x}_{t|t} &= \hat{x}_{t|t-1} + \mathbf{K}_t (z_t - \mathbf{H}_t \hat{x}_{t|t-1}), \\
    \mathbf{P}_{t|t} &= (\mathbf{I} - \mathbf{K}_t \mathbf{H}_t) \mathbf{P}_{t|t-1},
\end{aligned}
\end{equation}
 where \(\mathbf{P}_{t|t-1}\) is the error covariance matrix, and \(\mathbf{R}_t\) is the observation noise covariance. \(\mathbf{K}_t\) weights the correction based on the observed measurement \(z_t\). This leads to an updated state estimate and reduced uncertainty in \(\mathbf{P}_{t|t}\).
 %
%
OC-SORT~\cite{cao2023observation} introduces Observation-Centric Re-Update (ORU) to enhance KF robustness and address scenarios with lost detections. ORU compensates for missed observations by re-updating KF parameters as follows:
\setlength{\abovedisplayskip}{5pt}   
\setlength{\belowdisplayskip}{5pt}   
\begin{equation}
\label{eq:kfreupdate}
\begin{aligned}
    \mathbf{K}_t &= \mathbf{P}_{t|t-1} \mathbf{H}_t^\top \left( \mathbf{H}_t \mathbf{P}_{t|t-1} \mathbf{H}_t^\top + \mathbf{R}_t \right)^{-1}, \\
    \hat{x}_{t|t} &= \hat{x}_{t|t-1} + \mathbf{K}_t (\tilde{z}_t - \mathbf{H}_t \hat{x}_{t|t-1}), \\
    \mathbf{P}_{t|t} &= (\mathbf{I} - \mathbf{K}_t \mathbf{H}_t) \mathbf{P}_{t|t-1},
\end{aligned}
\end{equation}
where \(\mathbf{\tilde{z}}_t\) is the trajectory interpolating between the last-seen observation \(\mathbf{\tilde{z}}_{t_1}\) and the re-associated observation \(\mathbf{\tilde{z}}_{t_2}\) using: 
\setlength{\abovedisplayskip}{5pt}   
\setlength{\belowdisplayskip}{5pt}   
\begin{equation}
\label{eq:trajckf}
\begin{aligned}
    \mathbf{\tilde{z}}_t &= \text{Traj}_{virt}(\mathbf{\tilde{z}}_{t_1}, \mathbf{\tilde{z}}_{t_2}, t), \quad t_1 < t < t_2
\end{aligned}
\end{equation}
where \(\text{Traj}_{virt}\) denotes the interpolation function, enabling backtracking over missing frames.
We define each KF state as \( x = [u,v,s,r,\dot{u},\dot{v},\dot{s}]^{T} \), where \((u,v)\) denote the 2D coordinates of the object's center, \(s\) the bounding box scale, and \(r\) the aspect ratio of the bounding box (assumed constant). The derivatives \(\dot{u}\), \(\dot{v}\), and \(\dot{s}\) correspond to the temporal changes in \(u\), \(v\), and \(s\), respectively~\cite{cao2023observation}. Additionally, each detected object provides a bounding box defined as \( d = [u,v,w,h,c]^{T} \), where \((u,v)\) are the center coordinates, \(w\) and \(h\) are the width and height, and \(c\) the detection confidence.

As shown in Eq.~\ref{eq:matchhas}, in the association step, our framework calculates a matching score matrix linking observations in the current frame to existing tracklets. This matrix includes two motion-based components: \(S_{\text{IoU}}\) and \(S_{\text{ang}}\). The score \(S_{\text{IoU}}\) is based on the IoU between bounding boxes, which measures their spatial overlap. In parallel, \(S_{\text{ang}}\) captures the directional similarity between the new observations and existing tracklets. Higher values for \(S_{\text{IoU}}\) and \(S_{\text{ang}}\) reflect greater similarity between the observation and the corresponding tracklet, while lower values suggest divergence~\cite{cao2023observation}.

\textbf{ReIdentification:} To extract appearance features, we utilize the pre-trained FastReID~\cite{he2023fastreid}, a CNN-based RE-ID model that has been trained for the MOT~\cite{maggiolino2023deep, lv2024diffmot}. FastReID extracts features from each cropped object in the frame, which are then compared across tracklets and observations using cosine similarity, yielding an appearance-based score matrix \(S_{\text{emb}}\). To maintain consistent appearance features for each tracklet, we apply an exponential moving average (EMA)~\cite{aharon2022bot, du2023strongsort}, updating each tracklet's embeddings after each association step.

The embedding update for each tracklet is given by:
\begin{equation}
\label{eq:embupdate}
    emb_t = \mathcal{C} \cdot emb_{t-1} + (1 - \mathcal{C}) \cdot emb_{new},
\end{equation}
where \(\mathcal{C}\) is a dynamic coefficient defined as in~\cite{maggiolino2023deep}, $\mathcal{C} = \mathcal{T} + (1 - \mathcal{T}) \cdot \left(1 - \frac{c - \text{thresh}}{1 - \text{thresh}}\right)$.
In this formulation, \(c\) represents the detection confidence, \(\text{thresh}\) is a detection threshold, and \(\mathcal{T}\) is a fixed parameter, set to \(0.95\) following prior work. 

\vspace*{-0.4cm}
\subsection{Promptable Visual Segmentation (PVS)}
\vspace*{-0.2cm}
\label{sec:pvsseg}
For each frame, the object detection outputs bounding boxes. To have a fine object shape alignment, we apply Promptable Visual Segmentation (PVS), which extends static image segmentation to the video domain by enabling the generation of a spatio-temporal mask for a segment of interest across frames. In PVS, a prompt, such as a point, bounding box, or initial mask, is applied to a frame to define the target object, and the model then propagates this mask through subsequent frames. This task has emerged as a powerful method for fine-grained object segmentation in video sequences~\cite{delatolas2024learning}.
For shape alignment, we integrate Segment Anything Model 2 (SAM2)~\cite{ravi2024sam}, an advanced PVS framework designed for images and videos. SAM2 has some tracking capabilities but is not designed to track objects by matching bounding boxes and ID continuity, as required in MOT. 
Our shape alignment approach works as follows:
1) Given the bounding boxes of tracklets (i.e., objects already being tracked) in the previous frame, SAM2 generates segmentation masks for these tracklets. These masks represent the precise object shape within the bounding boxes.
2) For objects newly detected in the current frame \(t\), SAM2 uses the bounding box of each detected object in frame \(t\) to propagate the segmentation backward to find the corresponding object mask in frame \(t-1\). This allows us to retrieve the segmentation of each newly detected object in the previous frame, even if that object was not explicitly tracked before.
3) Once we have segmentation masks for both the existing tracklets (from frame \(t-1\)) and the newly detected objects (propagated backward from frame \(t\)), we perform mask matching within frame \(t-1\) by computing the mask IoU between each tracklet’s mask and the mask of each newly detected object. More details on how these IoUs are integrated into our association process can be found in Sec.~\ref{sec:HAS}.
By incorporating this segmentation-based mask matching, our method ensures that newly detected objects in frame \(t\) are correctly linked to their corresponding tracklets from frame \(t-1\), thereby improving tracking consistency and reducing ID switches. 
\vspace*{-0.4cm}
\subsection{Zero-Shot Depth Estimation}
\vspace*{-0.2cm}
\label{sec:depth}

To enrich our MOT framework with 3D spatial context, we leverage zero-shot depth estimation to capture object distances from the camera. Conventional zero-shot depth models have often relied on camera-specific metadata, such as intrinsics, which can be unavailable or unreliable. Recent methods, such as ZeroDepth~\cite{guizilini2023towards} and DMD~\cite{saxena2023zero}, perform well in predicting depth maps. However, these methods are limited by their dependence on precise camera parameters, which reduces generalizability. Depth Pro~\cite{bochkovskii2024depth} overcomes this limitation by achieving accurate depth estimation without requiring intrinsics. With its ability to estimate depth at high speed and accuracy, as well as to predict focal length from a single image, Depth Pro is ideally suited for MOT applications where efficiency and generalization are critical.

We process each frame in the video sequence using Depth Pro, which produces a dense depth map of the scene. This map provides a relative spatial representation, allowing us to measure the depth of each object by isolating the depth values within its bounding box. Depth-based measurements add a valuable discriminative layer by distinguishing visually similar objects that occupy different spatial planes, improving tracking accuracy in crowded scenarios.

We found that histogram-based vectorizations of depth values yielded the most effective results. By constructing a histogram of depth values within each object’s bounding box, we obtain a compact yet powerful descriptor of the object's 3D characteristics. To determine the depth similarity between frames, we compute the cosine similarity between these histograms, quantifying the similarity of spatial distributions.

As shown in Eq.~\ref{eq:matchhas}, we integrate the resulting depth similarity score into the overall matching score matrix. Fig.~\ref{fig:model_architecture} (a) illustrates this approach with two objects in the same frame, where histogram differences highlight the depth variance corresponding to their distance from the camera. This example demonstrates the added discriminatory power depth brings to object tracking. Fig.~\ref{fig:model_architecture} (b) and (c) show examples of zero-shot depth estimation and PVS modules.
\vspace*{-0.4cm}
\subsection{Hierarchical Alignment Score (HAS)}
\vspace*{-0.2cm}
\label{sec:HAS}

IoU is widely used in MOT to measure spatial overlap and associate objects across consecutive frames. Traditional IoU measures the overlap between bounding boxes and ranges from 0 to 1, capturing spatial alignment but overlooking important shape-based cues. Consequently, purely motion-based methods often fall short in complex scenes with occlusions, where bounding boxes alone may not accurately represent object shape or position. While recent appearance-based approaches~\cite{aharon2022bot, maggiolino2023deep} incorporate visual features, these methods only extract appearance from cropped bounding boxes, neglecting the spatial context within each frame. 

To address these limitations, we introduce a novel matching metric, HAS, which combines coarse alignment via bounding box IoU with fine object shape alignment via pixel-level IoU, progressively refining object matching. HAS enhances the matching robustness by integrating both spatial and shape information, ensuring that matches are not only spatially aligned but also consistent in shape alignment.

We calculate the HAS score \(S_{\text{HAS}}\) as follows:
\begin{equation}
\label{eq:hasss}
    S_{\text{HAS}}(\hat{X}, D) = S_{\text{IoU}_{bbox}}(\hat{X}, D) \times exp^{S_{\text{IoU}_{Seg}}(\hat{X}, D),}
\end{equation}
where \(\hat{X}\) denotes the predicted state of a tracklet, including its bounding box or segmentation mask estimated in the current frame, and \(D\) represents a new detection in the current frame, including the observed bounding box or segmentation mask. \(S_{\text{IoU}_{bbox}}\) represents the bounding box IoU, facilitating initial coarse matching based on location and scale, while \(S_{\text{IoU}_{Seg}}\) denotes the pixel-wise IoU, capturing precise shape alignment between objects. The exponential weighting on \(S_{\text{IoU}_{Seg}}\) ensures that minor improvements in shape alignment lead to significant increases in the HAS score, reinforcing high-fidelity matches.

\begin{figure}[!ht]
  \vspace*{-0.4cm}
  \centering
   \includegraphics[width=0.8\linewidth]{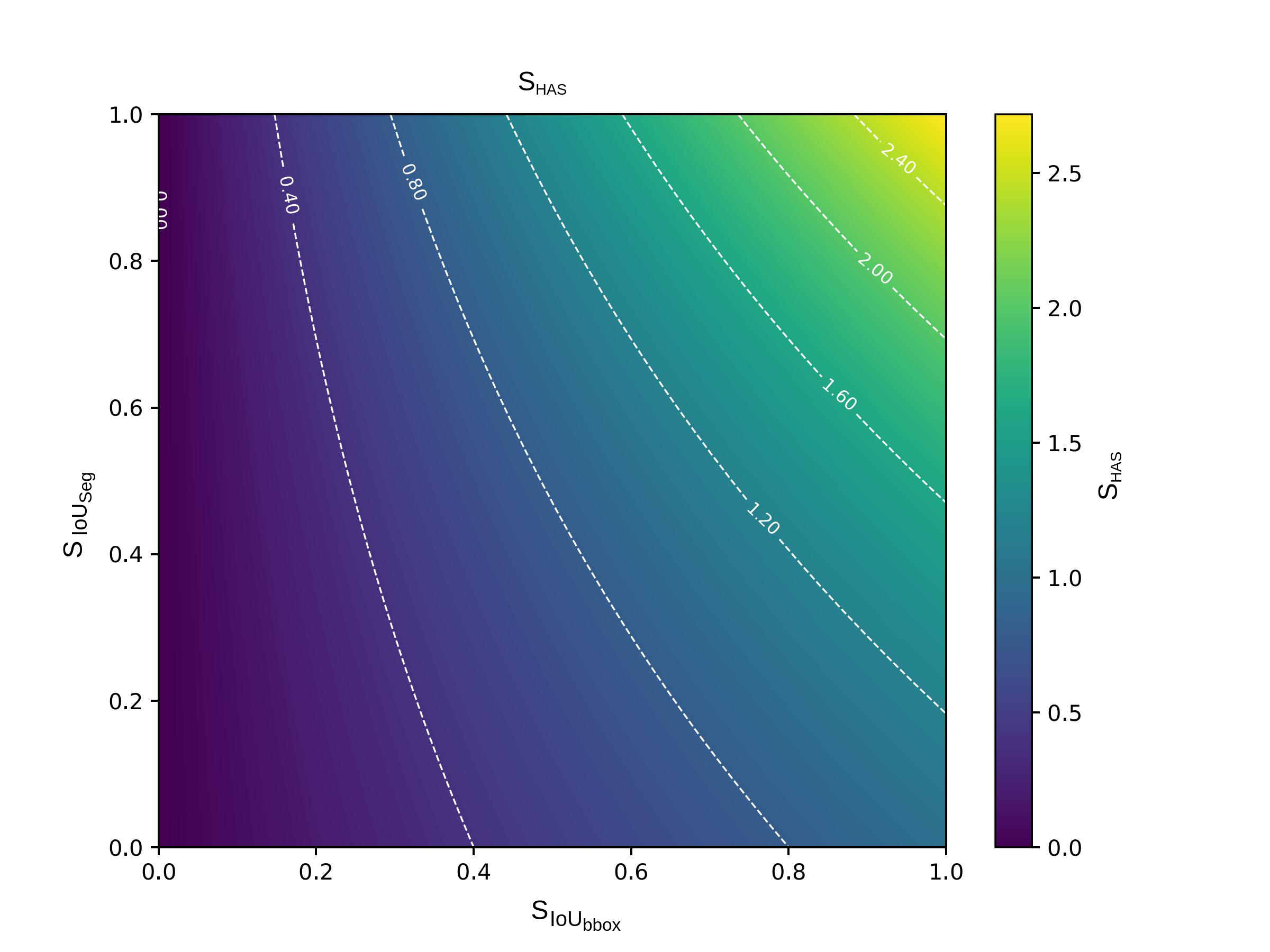}
   \vspace*{-0.4cm}
   \caption{Hierarchical Alignment Score (HAS). Heatmap of \(S_{\text{HAS}}\) as a function of bounding box IoU (\(S_{\text{IoU}_{bbox}}\)) and segment IoU (\(S_{\text{IoU}_{Seg}}\)). Contours indicate score levels, demonstrating the hierarchical influence of both spatial and shape alignment on the overall score. A high \(S_{\text{HAS}}\) suggests high similarity, while a low \(S_{\text{HAS}}\) indicate dissimilarity}
   \vspace*{-0.4cm}
   \label{fig:HAS_IoU_shape}
\end{figure}

The core innovation of HAS lies in its hierarchical alignment process. During the initial stages, the matching function prioritizes bounding box IoU, yielding a coarse spatial alignment based on object location and scale. As bounding box alignment improves, the influence of the exponential pixel-wise IoU term \(exp^{S_{\text{IoU}_{Seg}}}\) becomes more pronounced, refining alignment based on the actual object shapes. This hierarchical refinement ensures robust association by balancing location and shape information, progressively emphasizing fine shape details as alignment quality increases. Fig.~\ref{fig:HAS_IoU_shape} illustrates a heatmap of \(S_{\text{HAS}}\), showing how it evolves through the initial stages as a function of \(S_{\text{IoU}_{bbox}}\) and \(S_{\text{IoU}_{Seg}}\); the exponential weighting sharply amplifies the influence of fine-grained segmentation early on, particularly when bounding box overlap is low, emphasizing precise matching.

\vspace*{-0.4cm}
\subsection{Total Matching Score and Linear Solver}
\label{sec:matching}
\vspace*{-0.2cm}
To further enhance tracking performance, we incorporate HAS into a comprehensive score function, which integrates motion, appearance, and depth cues as follows:
\begin{equation}
    \label{eq:matchhas}
    \begin{aligned}
        Match_t = & \; S_{\text{HAS}_{t}}(\hat{X}, D) + S_{\text{ang}_{t}}(\hat{X}, D) \\
        & + S_{\text{depth}_{t}}(\hat{X}, D) + S_{\text{emb}_{t}}(\hat{X}, D),
    \end{aligned}
\end{equation}
where \(S_{\text{ang}_{t}}(\hat{X}, D)\) captures motion direction alignment~\cite{maggiolino2023deep}, \(S_{\text{emb}_{t}}(\hat{X}, D)\) the appearance score, \(S_{\text{depth}_{t}}(\hat{X}, D)\) depth similarity, and \(S_{\text{HAS}_{t}}(\hat{X}, D)\) the proposed HAS score. This holistic approach creates a robust multi-cue framework that effectively balances motion, 2D and 3D spatial alignment, and appearance features. The matching score matrix \(Match_t\) in Eq.~\ref{eq:matchhas} is then negated, transforming it into a cost matrix compatible with the linear solver~\cite{bewley2016simple} for optimal assignment.

\vspace*{-0.4cm}
\section{Experimental Results}
\vspace*{-0.2cm}
%
Main performance metrics for MOT are Higher Order Tracking Accuracy (HOTA), ID-based F1 Score (IDF1), and Association Accuracy (AssA)~\cite{luiten2021hota}. HOTA assesses both detection and association accuracy. IDF1 and AssA primarily evaluate association performance. Multi-Object Tracking Accuracy (MOTA) metric focuses on detection accuracy. In our results tables, \textuparrow means the higher, the better, and \textdownarrow means the lower, the better. We adopt YOLOX~\cite{zheng2021yolox} as the object detector. In our tables, MOT methods using YOLOX as the detector are highlighted in blue and bold indicates the best performance among TBD trackers (bottom part of each table).

\vspace*{-0.4cm}
\subsection{Benchmark Evaluation}
\vspace*{-0.2cm}
\noindent \textbf{DanceTrack:}  
This dataset includes videos with non-linear motion, frequent occlusions, and crossovers. As seen in Table~\ref{tab:dancetrack_comparison}, compared to the state-of-the-art TBD tracker DiffMOT~\cite{lv2024diffmot}, which requires training a computationally intensive transformer model for each dataset, DepthMOT achieves superior results without training for specific datasets. Our framework improves HOTA by 1.97\%, AssA by 3.84\%, and IDF1 by 3.54\% over DiffMOT, highlighting the effectiveness of our HAS and depth.
\footnotesize
\begin{table}[ht]
  \footnotesize
  \centering
  \setlength{\tabcolsep}{0.5pt} 
  \renewcommand{\arraystretch}{0.9} 
  \resizebox{0.8\columnwidth}{!}{%
    \begin{tabular}{@{}lccccc@{}}
      \toprule
      Method & HOTA\textuparrow & IDF1\textuparrow & AssA\textuparrow & MOTA\textuparrow & DetA\textuparrow \\
      \midrule
      FairMOT~\cite{zhang2021fairmot} & 39.7 & 40.8 & 23.8 & 82.2 & 66.7 \\
      CenterTrack~\cite{zhou2020tracking} & 41.8 & 35.7 & 22.6 & 86.8 & 78.1 \\
      TraDes~\cite{wu2021track} & 43.3 & 41.2 & 25.4 & 86.2 & 74.5 \\
      TransTrack~\cite{sun2020transtrack} & 45.5 & 45.2 & 27.5 & 85.4 & 75.9 \\
      DiffusionTrack~\cite{luo2024diffusiontrack} & 52.4 & 47.5 & 33.5 & 89.5 & 82.2 \\
      MeMOTR~\cite{gao2023memotr} & 68.5 & 71.2 & 58.4 & 89.9 & 80.5 \\
      \rowcolor{lightblue} MOTRv2~\cite{zhang2023motrv2} & 69.9 & 71.7 & 59.0 & 91.9 & 83.0 \\
      \midrule
      \rowcolor{lightblue} GHOST~\cite{seidenschwarz2023simple} & 56.7 & 57.7 & 39.8 & 91.3 & 81.1 \\
      \rowcolor{lightblue} ByteTrack~\cite{zhang2022bytetrack} & 47.3 & 52.5 & 31.4 & 89.5 & 71.6 \\
      \rowcolor{lightblue} SORT~\cite{bewley2016simple} & 47.9 & 50.8 & 31.2 & 91.8 & 72.0 \\
      \rowcolor{lightblue} MotionTrack~\cite{xiao2024motiontrack} & 52.9 & 53.8 & 34.7 & 91.3 & 80.9 \\
      \rowcolor{lightblue} OC-SORT~\cite{cao2023observation} & 55.1 & 54.2 & 38.0 & 89.4 & 80.3 \\
      \rowcolor{lightblue} StrongSORT++~\cite{du2023strongsort} & 55.6 & 55.2 & 38.6 & 91.1 & 80.7 \\
      \rowcolor{lightblue} GeneralTrack~\cite{qin2024towards} & 59.2 & 59.7 & 42.8 & 91.8 & 82.0 \\
      \rowcolor{lightblue} C-BIoU~\cite{yang2023hard} & 60.6 & 61.6 & 45.4 & 91.8 & 81.3 \\
      \rowcolor{lightblue} Deep OC-SORT~\cite{maggiolino2023deep} & 61.3 & 61.5 & 45.8 & 92.3 & 82.2 \\
      \rowcolor{lightblue} CMTrack~\cite{shim2024confidence} & 61.8 & 63.3 & 46.4 & 92.5 & - \\
      \rowcolor{lightblue} DiffMOT~\cite{lv2024diffmot} & 62.3 & 63.0 & 47.2 & \textbf{92.8} & \textbf{82.5} \\
      \rowcolor{lightblue} DepthMOT & \textbf{64.27} & \textbf{66.54} & \textbf{51.04} & 90.08 & 81.07 \\
      \bottomrule
    \end{tabular}%
  }
  \caption{Comparison of JDR and TBD trackers on the DanceTrack \textbf{test set}. Our DepthMOT outperforms the state-of-the-art TBD tracker DiffMOT, which requires training on each dataset separately.}
  \label{tab:dancetrack_comparison}
\end{table}
\normalsize
\tiny
\begin{table}[!ht]
  \tiny
  \centering
  \setlength{\tabcolsep}{0.5pt}
  \renewcommand{\arraystretch}{0.9} 
  \resizebox{0.8\columnwidth}{!}{%
  \begin{tabular}{@{}lccccc@{}}
    \toprule
    Method & HOTA\textuparrow & IDF1\textuparrow & AssA\textuparrow & MOTA\textuparrow & DetA\textuparrow \\
    \midrule
    FairMOT~\cite{zhang2021fairmot} & 49.3 & 53.5 & 34.7 & 86.4 & 70.2 \\
    GTR~\cite{zhou2022global} & 54.5 & 55.8 & 45.9 & 67.9 & 64.8 \\
    CenterTrack~\cite{zhou2020tracking} & 62.7 & 60.0 & 48.0 & 90.8 & 81.7 \\
    TransTrack~\cite{sun2020transtrack} & 68.9 & 71.5 & 57.5 & 92.6 & 82.7 \\
    \midrule
    \rowcolor{lightblue} ByteTrack~\cite{zhang2022bytetrack} & 62.8 & 69.8 & 51.2 & 94.1 & 77.1 \\
    \rowcolor{lightblue} BoT-SORT~\cite{aharon2022bot} & 68.7 & 70.0 & 55.9 & 94.5 & 84.4 \\
    \rowcolor{lightblue} OC-SORT~\cite{cao2023observation} & 71.9 & 72.2 & 59.8 & 94.5 & 86.4 \\
    \rowcolor{lightblue} *ByteTrack~\cite{zhang2022bytetrack} & 64.1 & 71.4 & 52.3 & 95.9 & 78.5 \\
    \rowcolor{lightblue} *MixSort-Byte~\cite{cui2023sportsmot} & 65.7 & 74.4 & 58.4 & 96.2 & 78.8 \\
    \rowcolor{lightblue} *OC-SORT~\cite{cao2023observation} & 73.7 & 74.0 & 61.5 & 96.5 & 88.5 \\
    \rowcolor{lightblue} *MixSort-OC~\cite{cui2023sportsmot} & 74.1 & 74.4 & 62.0 & 96.5 & 88.5 \\
    \rowcolor{lightblue} *GeneralTrack~\cite{qin2024towards} & 74.1 & 76.4 & 61.7 & 96.8 & 89.0 \\
    \rowcolor{lightblue} *DiffMOT~\cite{lv2024diffmot} & \textbf{76.2} & 76.1 & 65.1 & \textbf{97.1} & \textbf{89.3} \\
    \rowcolor{lightblue} *DepthMOT & \textbf{76.2} & \textbf{76.9} & \textbf{65.6} & 95.9 & 88.5 \\
    \bottomrule
  \end{tabular}
  }
  \caption{Comparison with JDR and TBD trackers on the SportsMOT \textbf{test set}. * indicates that the detector is trained on the SportsMOT train and validation sets.}
  \label{tab:sportmot_comparison}
  \vspace*{-0.3cm}
\end{table}
\normalsize

\noindent \textbf{SportsMOT:}  
It involves fast and unpredictable movements, including frequent interaction, similar objects, and occlusion. Compared to the state-of-the-art DiffMOT~\cite{lv2024diffmot}, our DepthMOT has comparable performance as seen in  Table~\ref{tab:sportmot_comparison}.

%
Under MOT17 and MOT20 datasets (linear motion), no tracker performs best overall (see supplementary material at \href{https://sigport.org/sites/all/modules/pubdlcnt/pubdlcnt.php?fid=10006}{https://sigport.org/sites/all/modules/pubdlcnt/pubdlcnt.php?
fid=10006}). Ours achieves the lowest false positive rate, demonstrating the effectiveness of HAS. Compared to the state-of-the-art JDR tracker MOTRv2~\cite{zhang2023motrv2}, our TBD method is more efficient (MOTRv2 demands 8 GPUs for multi-stage training) and outperforms MOTRv2 under linear motion datasets  MOT17 and MOT20.
\vspace*{-0.4cm}
\subsection{Ablations}
\vspace*{-0.2cm}
We conduct an ablation study on the DanceTrack validation set to evaluate the impact of: the proposed Hierarchical Alignment Score (HAS) and the integration of depth. Table~\ref{tab:depthandHAS} presents different framework configurations with or without depth scores and HAS. As shown, the combination of both HAS and depth yields the highest accuracy, achieving $61.81\%$ HOTA, $49.12\%$ AssA, and $64.13\%$ IDF1. 
This improvement demonstrates that depth, as an additional spatial cue, is effective in enhancing object association.
\begin{table}[!ht]
  \centering
  \resizebox{0.9\columnwidth}{!}{%
  \begin{tabular}{@{}ccccc|ccc@{}}
    \toprule
    Appearance & Mask IoU & Bbox IoU & HAS & Depth & HOTA $\uparrow$ & AssA $\uparrow$ & IDF1 $\uparrow$ \\
    \midrule
    \checkmark & \checkmark & & &  & 54.78 & 38.54 & 52.70 \\
    \checkmark & & \checkmark & &  & 59.46 & 43.92 & 59.08 \\
    \checkmark & & & \checkmark &  & 60.62 & 47.15 & 62.16 \\
    \checkmark & & \checkmark &  & \checkmark & 60.45 & 47.02 & 62.13 \\
    \checkmark & & & \checkmark & \checkmark & \textbf{61.81} & \textbf{49.12} & \textbf{64.13} \\
    \bottomrule
  \end{tabular}%
  }
  \caption{Ablation study: Impact of HAS and depth on DanceTrack \textbf{validation set} performance.}
  \label{tab:depthandHAS}
\end{table}
\vspace*{-0.4cm}
\section{CONCLUSION}
\vspace*{-0.2cm}
We proposed the DepthMOT framework for MOT, which incorporates depth information alongside visual and motion features. Additionally, we introduced a hierarchical alignment score (HAS) that prioritizes bounding boxes during the initial association phase and progressively balances pixel-level segments as matching improves. DepthMOT achieves state-of-the-art results on DanceTrack and SportsMOT, handling rapid non-linear object motion, high appearance similarity, frequent occlusions, and crossovers. Notably, DepthMOT achieves competitive performance using only pre-trained models without any training nor fine-tuning.


\clearpage

\footnotesize
\bibliographystyle{IEEEtran}
\bibliography{strings}

\begin{thebibliography}{10}
\providecommand{\url}[1]{#1}
\csname url@samestyle\endcsname
\providecommand{\newblock}{\relax}
\providecommand{\bibinfo}[2]{#2}
\providecommand{\BIBentrySTDinterwordspacing}{\spaceskip=0pt\relax}
\providecommand{\BIBentryALTinterwordstretchfactor}{4}
\providecommand{\BIBentryALTinterwordspacing}{\spaceskip=\fontdimen2\font plus
\BIBentryALTinterwordstretchfactor\fontdimen3\font minus \fontdimen4\font\relax}
\providecommand{\BIBforeignlanguage}[2]{{%
\expandafter\ifx\csname l@#1\endcsname\relax
\typeout{** WARNING: IEEEtran.bst: No hyphenation pattern has been}%
\typeout{** loaded for the language `#1'. Using the pattern for}%
\typeout{** the default language instead.}%
\else
\language=\csname l@#1\endcsname
\fi
#2}}
\providecommand{\BIBdecl}{\relax}
\BIBdecl

\bibitem{vaquero2024lost}
L.~Vaquero, Y.~Xu, X.~Alameda-Pineda, V.~M. Brea, and M.~Mucientes, ``Lost and {F}ound: {O}vercoming {D}etector {F}ailures in {O}nline {M}ulti-object {T}racking,'' in \emph{Eur. Conf. Comput. Vis.}\hskip 1em plus 0.5em minus 0.4em\relax Springer, 2024, pp. 448--466.

\bibitem{lv2024diffmot}
W.~Lv, Y.~Huang, N.~Zhang, R.-S. Lin, M.~Han, and D.~Zeng, ``Diff{M}ot: A {R}eal-time {D}iffusion-based {M}ultiple {O}bject {T}racker with {N}on-linear {P}rediction,'' in \emph{IEEE Conf. Comput. Vis. Pattern Recog.}, 2024, pp. 19\,321--19\,330.

\bibitem{shim2024confidence}
K.~Shim, J.~Hwang, K.~Ko, and C.~Kim, ``A {C}onfidence-{A}ware {M}atching {S}trategy for {G}eneralized {M}ulti-{O}bject {T}racking,'' in \emph{IEEE Int. Conf. Image Process.}\hskip 1em plus 0.5em minus 0.4em\relax IEEE, 2024, pp. 4042--4048.

\bibitem{zhang2021fairmot}
Y.~Zhang, C.~Wang, X.~Wang, W.~Zeng, and W.~Liu, ``Fair{M}ot: {O}n the {F}airness of {D}etection and {R}e-identification in {M}ultiple {O}bject {T}racking,'' \emph{Int. J. Comput. Vis.}, vol. 129, pp. 3069--3087, 2021.

\bibitem{xu2022transcenter}
Y.~Xu, Y.~Ban, G.~Delorme, C.~Gan, D.~Rus, and X.~Alameda-Pineda, ``Trans{C}enter: {T}ransformers with {D}ense {R}epresentations for {M}ultiple-{O}bject {T}racking,'' \emph{IEEE Trans. Pattern Anal. Mach. Intell.}, vol.~45, no.~6, pp. 7820--7835, 2022.

\bibitem{cao2023observation}
J.~Cao, J.~Pang, X.~Weng, R.~Khirodkar, and K.~Kitani, ``Observation-{C}entric {SORT}: {R}ethinking {SORT} for {R}obust {M}ulti-{O}bject {T}racking,'' in \emph{IEEE Conf. Comput. Vis. Pattern Recog.}, 2023, pp. 9686--9696.

\bibitem{maggiolino2023deep}
G.~Maggiolino, A.~Ahmad, J.~Cao, and K.~Kitani, ``Deep {OC-SORT}: Multi-{P}edestrian {T}racking by {A}daptive {R}e-{I}dentification,'' in \emph{IEEE Int. Conf. Image Process.}\hskip 1em plus 0.5em minus 0.4em\relax IEEE, 2023, pp. 3025--3029.

\bibitem{meinhardt2022trackformer}
T.~Meinhardt, A.~Kirillov, L.~Leal-Taixe, and C.~Feichtenhofer, ``Trackformer: {M}ulti-{O}bject {T}racking with {T}ransformers,'' in \emph{IEEE Conf. Comput. Vis. Pattern Recog.}, 2022, pp. 8844--8854.

\bibitem{braso2020learning}
G.~Bras{\'o} and L.~Leal-Taix{\'e}, ``Learning a {N}eural {S}olver for {M}ultiple {O}bject {T}racking,'' in \emph{IEEE Conf. Comput. Vis. Pattern Recog.}, 2020, pp. 6247--6257.

\bibitem{cetintas2023unifying}
O.~Cetintas, G.~Bras{\'o}, and L.~Leal-Taix{\'e}, ``Unifying {S}hort and {L}ong-{T}erm {T}racking with {G}raph {H}ierarchies,'' in \emph{IEEE Conf. Comput. Vis. Pattern Recog.}, 2023, pp. 22\,877--22\,887.

\bibitem{braso2022multi}
G.~Bras{\'o}, O.~Cetintas, and L.~Leal-Taix{\'e}, ``Multi-{O}bject {T}racking and {S}egmentation {V}ia {N}eural {M}essage {P}assing,'' \emph{Int. J. Comput. Vis.}, vol. 130, no.~12, pp. 3035--3053, 2022.

\bibitem{zhang2023motrv2}
Y.~Zhang, T.~Wang, and X.~Zhang, ``{MOTR}v2: {B}ootstrapping {E}nd-to-{E}nd {M}ulti-{O}bject {T}racking by {P}retrained {O}bject {D}etectors,'' in \emph{IEEE Conf. Comput. Vis. Pattern Recog.}, 2023, pp. 22\,056--22\,065.

\bibitem{zheng2021yolox}
G.~Zheng, L.~Songtao, W.~Feng, L.~Zeming, S.~Jian \emph{et~al.}, ``Yolox: Exceeding yolo series in 2021,'' \emph{arXiv preprint arXiv:2107.08430}, 2021.

\bibitem{gao2023memotr}
R.~Gao and L.~Wang, ``Me{MOTR}: {L}ong-{T}erm {M}emory-{A}ugmented {T}ransformer for {M}ulti-{O}bject {T}racking,'' in \emph{Int. Conf. Comput. Vis.}, 2023, pp. 9901--9910.

\bibitem{huang2024deconfusetrack}
C.~Huang, S.~Han, M.~He, W.~Zheng, and Y.~Wei, ``Deconfuse{T}rack: {D}ealing with {C}onfusion for {M}ulti-{O}bject {T}racking,'' in \emph{IEEE Conf. Comput. Vis. Pattern Recog.}, 2024, pp. 19\,290--19\,299.

\bibitem{quach2024depth}
K.~G. Quach, P.~Nguyen, C.~N. Duong, T.~D. Bui, and K.~Luu, ``{D}epth {P}erspective-{A}ware {M}ultiple {O}bject {T}racking,'' in \emph{Eng. Appl. AI Swarm Intell.}\hskip 1em plus 0.5em minus 0.4em\relax Springer, 2024, pp. 181--205.

\bibitem{wang2024robust}
J.~Wang, H.~Zheng, Y.~Yu, Y.~He, and Y.~Liu, ``Robust multiple obstacle tracking method based on depth aware {OCSORT} for agricultural robots,'' \emph{Comput. Electron. Agric.}, vol. 217, p. 108580, 2024.

\bibitem{liu2022det}
C.-J. Liu and T.-N. Lin, ``{DET}: {D}epth-{E}nhanced {T}racker to {M}itigate {S}evere {O}cclusion and {H}omogeneous {A}ppearance {P}roblems for {I}ndoor {M}ultiple-{O}bject {T}racking,'' \emph{IEEE Access}, vol.~10, pp. 8287--8304, 2022.

\bibitem{zhang2022bytetrack}
Y.~Zhang, P.~Sun, Y.~Jiang, D.~Yu, F.~Weng, Z.~Yuan, P.~Luo, W.~Liu, and X.~Wang, ``{B}yte{T}rack: {M}ulti-object {T}racking by {A}ssociating {E}very {D}etection {B}ox,'' in \emph{Eur. Conf. Comput. Vis.}\hskip 1em plus 0.5em minus 0.4em\relax Springer, 2022, pp. 1--21.

\bibitem{he2023fastreid}
L.~He, X.~Liao, W.~Liu, X.~Liu, P.~Cheng, and T.~Mei, ``Fastreid: A {P}ytorch {T}oolbox for {G}eneral {I}nstance {R}e-identification,'' in \emph{ACM Int. Conf. Multimedia}, 2023, pp. 9664--9667.

\bibitem{aharon2022bot}
N.~Aharon, R.~Orfaig, and B.-Z. Bobrovsky, ``{B}o{T}-{SORT}: {R}obust {A}ssociations {M}ulti-{P}edestrian {T}racking,'' \emph{arXiv preprint arXiv:2206.14651}, 2022.

\bibitem{du2023strongsort}
Y.~Du, Z.~Zhao, Y.~Song, Y.~Zhao, F.~Su, T.~Gong, and H.~Meng, ``Strong{SORT}: {M}ake {D}eep{SORT} {G}reat {A}gain,'' \emph{IEEE Trans. Multimedia}, vol.~25, pp. 8725--8737, 2023.

\bibitem{delatolas2024learning}
T.~Delatolas, V.~Kalogeiton, and D.~P. Papadopoulos, ``{L}earning the {W}hat and {H}ow of {A}nnotation in {V}ideo {O}bject {S}egmentation,'' in \emph{IEEE Winter Conf. Appl. Comput. Vis.}, 2024, pp. 6951--6961.

\bibitem{ravi2024sam}
N.~Ravi, V.~Gabeur, Y.-T. Hu, R.~Hu, C.~Ryali, T.~Ma, H.~Khedr, R.~R{\"a}dle, C.~Rolland, L.~Gustafson \emph{et~al.}, ``{SAM} 2: Segment {A}nything in {I}mages and {V}ideos,'' \emph{arXiv preprint arXiv:2408.00714}, 2024.

\bibitem{guizilini2023towards}
V.~Guizilini, I.~Vasiljevic, D.~Chen, R.~Ambruș, and A.~Gaidon, ``{T}owards {Z}ero-{S}hot {S}cale-{A}ware {M}onocular {D}epth {E}stimation,'' in \emph{IEEE Conf. Comput. Vis. Pattern Recog.}, 2023, pp. 9233--9243.

\bibitem{saxena2023zero}
S.~Saxena, J.~Hur, C.~Herrmann, D.~Sun, and D.~J. Fleet, ``{Z}ero-{S}hot {M}etric {D}epth with a {F}ield-of-{V}iew {C}onditioned {D}iffusion {M}odel,'' \emph{arXiv preprint arXiv:2312.13252}, 2023.

\bibitem{bochkovskii2024depth}
A.~Bochkovskii, A.~Delaunoy, H.~Germain, M.~Santos, Y.~Zhou, S.~R. Richter, and V.~Koltun, ``Depth {P}ro: {S}harp {M}onocular {M}etric {D}epth in {L}ess {T}han a {S}econd,'' \emph{arXiv preprint arXiv:2410.02073}, 2024.

\bibitem{bewley2016simple}
A.~Bewley, Z.~Ge, L.~Ott, F.~Ramos, and B.~Upcroft, ``Simple online and realtime tracking,'' in \emph{IEEE Int. Conf. Image Process.}\hskip 1em plus 0.5em minus 0.4em\relax IEEE, 2016, pp. 3464--3468.

\bibitem{luiten2021hota}
J.~Luiten, A.~Osep, P.~Dendorfer, P.~Torr, A.~Geiger, L.~Leal-Taix{\'e}, and B.~Leibe, ``{HOTA}: A {H}igher {O}rder {M}etric for {E}valuating {M}ulti-object {T}racking,'' \emph{Int. J. Comput. Vis.}, vol. 129, pp. 548--578, 2021.

\bibitem{zhou2020tracking}
X.~Zhou, V.~Koltun, and P.~Kr{\"a}henb{\"u}hl, ``{T}racking {O}bjects as {P}oints,'' in \emph{Eur. Conf. Comput. Vis.}\hskip 1em plus 0.5em minus 0.4em\relax Springer, 2020, pp. 474--490.

\bibitem{wu2021track}
J.~Wu, J.~Cao, L.~Song, Y.~Wang, M.~Yang, and J.~Yuan, ``{T}rack {T}o {D}etect and {S}egment: {A}n {O}nline {M}ulti-{O}bject {T}racker,'' in \emph{IEEE Conf. Comput. Vis. Pattern Recog.}, 2021, pp. 12\,352--12\,361.

\bibitem{sun2020transtrack}
P.~Sun, J.~Cao, Y.~Jiang, R.~Zhang, E.~Xie, Z.~Yuan, C.~Wang, and P.~Luo, ``Trans{T}rack: {M}ultiple {O}bject {T}racking with {T}ransformer,'' \emph{arXiv preprint arXiv:2012.15460}, 2020.

\bibitem{luo2024diffusiontrack}
R.~Luo, Z.~Song, L.~Ma, J.~Wei, W.~Yang, and M.~Yang, ``{D}iffusion{T}rack: {D}iffusion {M}odel for {M}ulti-{O}bject {T}racking,'' in \emph{AAAI}, vol.~38, no.~5, 2024, pp. 3991--3999.

\bibitem{seidenschwarz2023simple}
J.~Seidenschwarz, G.~Bras{\'o}, V.~C. Serrano, I.~Elezi, and L.~Leal-Taix{\'e}, ``{S}imple {C}ues {L}ead to a {S}trong {M}ulti-{O}bject {T}racker,'' in \emph{IEEE Conf. Comput. Vis. Pattern Recog.}, 2023, pp. 13\,813--13\,823.

\bibitem{xiao2024motiontrack}
C.~Xiao, Q.~Cao, Y.~Zhong, L.~Lan, X.~Zhang, Z.~Luo, and D.~Tao, ``{M}otion{T}rack: {L}earning motion predictor for multiple object tracking,'' \emph{Neural Networks}, vol. 179, p. 106539, 2024.

\bibitem{qin2024towards}
Z.~Qin, L.~Wang, S.~Zhou, P.~Fu, G.~Hua, and W.~Tang, ``{T}owards {G}eneralizable {M}ulti-{O}bject {T}racking,'' in \emph{IEEE Conf. Comput. Vis. Pattern Recog.}, 2024, pp. 18\,995--19\,004.

\bibitem{yang2023hard}
F.~Yang, S.~Odashima, S.~Masui, and S.~Jiang, ``{H}ard to {T}rack {O}bjects {W}ith {I}rregular {M}otions and {S}imilar {A}ppearances? {M}ake {I}t {E}asier by {B}uffering the {M}atching {S}pace,'' in \emph{IEEE Winter Conf. Appl. Comput. Vis.}, 2023, pp. 4799--4808.

\bibitem{zhou2022global}
X.~Zhou, T.~Yin, V.~Koltun, and P.~Kr{\"a}henb{\"u}hl, ``{G}lobal {T}racking {T}ransformers,'' in \emph{IEEE Conf. Comput. Vis. Pattern Recog.}, 2022, pp. 8771--8780.

\bibitem{cui2023sportsmot}
Y.~Cui, C.~Zeng, X.~Zhao, Y.~Yang, G.~Wu, and L.~Wang, ``{S}ports{MOT}: {A} {L}arge {M}ulti-{O}bject {T}racking {D}ataset in {M}ultiple {S}ports {S}cenes,'' in \emph{Int. Conf. Comput. Vis.}, 2023, pp. 9921--9931.

\end{thebibliography}
\normalsize

\end{document}